\documentclass[conference]{IEEEtran}
\IEEEoverridecommandlockouts
\usepackage{cite}
\usepackage{amsmath,amssymb,amsfonts}
\usepackage{algorithmic}
\usepackage{graphicx}
\usepackage{textcomp}
\usepackage{xcolor}

\usepackage{color}
\usepackage{csquotes}
\usepackage{enumerate}
\usepackage{dsfont} 
\usepackage{amsfonts}
\usepackage{graphicx}
\usepackage{amsmath} 
\usepackage{bm}
\usepackage{booktabs}
\usepackage{float}
\usepackage{siunitx}
\usepackage{multirow}
\usepackage{amsthm}
\usepackage{caption}
\usepackage{subcaption}
\usepackage{verbatim}
\newtheorem{thm}{Theorem}
\newtheorem{prp}{Proposition}[section]
\newtheorem{dfn}[prp]{Definition}

\newtheorem{conjecture}{Conjecture}	
\newtheorem{property}{Property}

\def\BibTeX{{\rm B\kern-.05em{\sc i\kern-.025em b}\kern-.08em
		T\kern-.1667em\lower.7ex\hbox{E}\kern-.125emX}}
\begin{document}
	
	\title{Implicit Regularization in Deep Tensor  Factorization
		\thanks{This work was funded by the French national research agency (grant number ANR16-CE23-0006 "Deep in France").}
	}
	
	\author{\IEEEauthorblockN{Paolo Milanesi$^{1}$, Hachem Kadri$^{1}$, Stéphane Ayache$^{1}$, Thierry Artières$^{1,2}$}
		\IEEEauthorblockA{$^{1}$\textit{Aix Marseille Univ, Université de Toulon, CNRS, LIS} \\
			$^{2}$\textit{Ecole Centrale de Marseille}\\
			Marseille, France  \\
			firstname.lastname@lis-lab.fr}
	}
	
	\maketitle
	
	\begin{abstract}
		Attempts of studying implicit regularization associated to gradient descent (GD) have identified matrix completion as  a suitable test-bed. Late findings suggest that this phenomenon cannot be phrased as a minimization-norm problem, implying that a paradigm shift is required and that dynamics has to be taken into account. In the present work we address the more general setup of tensor completion by leveraging two popularized tensor factorization, namely Tucker and Tensor\-Train (TT). We track relevant quantities such as tensor nuclear norm, \textit{effective rank}, generalized singular values and we introduce deep Tucker and TT unconstrained factorization to deal with the completion task. Experiments on both synthetic and real data show that gradient descent promotes solution with low-rank, and validate the conjecture saying that the phenomenon has to be addressed from a dynamical perspective.
	\end{abstract}
	
	
	\section{INTRODUCTION}
	
	Deep Neural Networks (DNNs) are currently the state of the art in a variety of tasks ranging from computer vision to speech recognition and bioinformatics. Though their deployment is ubiquitous,  a clear theoretical understanding still lies out of our reach; for example, currently employed DNNs are largely overparametrized and yet, they are able to generalize to unseen data \cite{Kaw, Zhang, neyshabur2017exploring}.
	Although a number of tricks (including small batch processing, early stopping of training and norms constraints over layers weights, just to name a few) are currently used by practitioners to enhance inference capacities, it has become clear that none of these devices is neither a sufficient nor a necessary condition to guarantee generalization of DNNs~\cite{Zhang}.
	Moreover, experiments show that DNNs trained by back propagation with vanilla gradient descent and no explicit regularization do generalize reasonably well \cite{Rud}, suggesting that the optimization algorithm plays a role in driving the loss function towards a local minimum with good generalization properties. Common wisdom says that implicit regularization induced by gradient descent can be rephrased by means of some variational principle: for example, it is known that for linear models, gradient descent finds the solution with minimal Frobenius norm \cite{Zhang}, thus one could expect this to hold true in more complex scenarios, possibly  by replacing Frobenius norm with other norms or quasi-norms architecture-depending.
	A recent line of research addresses this issue within the framework of matrix completion. To put it more clearly, let us consider a ground truth matrix $M\in\mathbb{R}^{n\times m}$ which is partially observed and let \break $\Omega \subset \{1, \ldots, n\}\times \{1, \ldots m\}$ be the set indexing the observed entries.
	The completion task consists in finding the matrix $W$ minimizing the loss
	\begin{equation}\label{matrix_completion_loss}
		\ell(W) = \frac{1}{2}\sum_{(i,j)\in\Omega}\left(W_{ij} - M_{ij}\right)^2.
	\end{equation}
	The above problem being ill-posed, one needs some additional constraints and the usual strategy consists in seeking for the minimizer with minimal rank, motivated by the observation that in many practical applications the sought matrix only has few degrees of freedom.
	However, as the rank minimization is a NP-hard problem~\cite{gillis2011low}, one is led to minimize the nuclear norm (\textit{i.e.} the sum of the singular values) which is a convex problem and formally equivalent \cite{Can}---provided that number of observed entries is large enough and under some ``incoherence'' assumption---to the rank minimization problem.
	A natural approach to solve this problem by means of gradient descent would be to factorize $W$ as $W=W_1W_2$, with $W_1\in\mathbb{R}^{n\times \hat{r}}$, $W_2\in\mathbb{R}^{\hat{r}\times m}$, with $\hat{r}$ some estimate of the rank of $M$, plug it in \eqref{matrix_completion_loss} and run the gradient descent algorithm. Notice that this corresponds to train a  $2$-layers Linear Neural Network on the dataset $\{(i,j), M_{ij}\}_{(i,j)\in\Omega}$ with loss given by Eq. \eqref{matrix_completion_loss}.
	
	However, a somewhat surprising finding is that,  even in the case where the shared dimension of $W_1, W_2$ is unconstrained, namely $\hat{r}=\min\{m, n\}$, the solution of~\eqref{matrix_completion_loss} via gradient descent turns out to be a low-rank matrix. This empirical evidence together with a proof in a restrictive setup led Gunasekar \textit{et al.} to conjecture the following:
	\begin{conjecture}[$\mkern-5mu$\cite{Gun}, roughly stated]\label{con1}
		With small enough~learning rate and initialization, gradient descent on full-dimensional matrix factorization converges to the solution with minimal nuclear norm.
	\end{conjecture}
	This conjecture supports the viewpoint of conventional learning theory, claiming that implicit regularization can be rephrased as a norm-minimization problem.
	
	Arora \textit{et al.} \cite{Arora} pushed the analysis further by considering deep matrix factorization, which consists in parametrizing $W$ in Eq. \eqref{matrix_completion_loss} as $W=W_LW_{L-1}\ldots W_2W_1$ for some $L\in\mathbb{N}$ and with $\{W_i\}_{i=1}^L$ to be such that no constraint on the rank is present. Notice that deep matrix factorization is a generalization of shallow matrix factorization setup investigated by \cite{Gun}, which corresponds to choose $L=2$. On the one hand they empirically showed that depth enhances recovery performances and, on the other hand, they found that the proof  supporting the  conjecture raised in \cite{Gun} carries over to the deep setup, meaning that the argument of \cite{Gun} cannot distinguish the shallow setup ($L=2$)  from the deep one. As this fact contradict experimental evidence
     and having proved that gradient descent over deep matrix factorization do not minimize any quasi-norm,
	they were led to challenge Conjecture \ref{con1} and to raise the following one:
	\begin{conjecture}[$\!\!$\cite{Arora}, roughly stated]\label{con2}
		Implicit regularization of gradient decent in deep matrix factorization cannot be formulated as a norm-minimization problem; instead one needs to take into account the trajectories of the gradient descent dynamics.
	\end{conjecture}
	In this respect they studied the gradient flow and they found out, both theoretically and empirically, that GD promotes sparsity of singular values of $W$.
	Lately,~\cite{Razin} exhibited a class of matrix completion problem for which fitting observed entries through gradient descent on deep matrix factorization leads all norms and quasi-norms to grow to infinity whereas other rank surrogates, such as effective rank (cf Section \ref{nuc_eff}), converge towards their minimum. This can be considered as one more clue to the fact that implicit regularization needs to be understood as a dynamical phenomenon.
	To summarize, not only an understanding of implicit regularization is lacking, but actually the mathematical formalism one needs to address the problem is unclear.
	Though the conventional viewpoint---claiming that the problem can be phrased as some norm minimization---is becoming more and more untrustworthy and  a shift of paradigm seems to be required, one cannot reject it yet.\par

	In the present work we address implicit regularization of gradient descent 
	in  the more general setup of tensor completion. Tensors can be considered as multi-dimensional arrays and they have been attracting an increasing interest as many problems require the manipulation of quantities which are, by their very nature, multivariates. Although tensor completion could be viewed as a byproduct of matrix completion and solved  by means of techniques developed for the latter, a more careful analysis shows that downgrading tensor completion to a  matrix framework---e.g. via slicings or unfoldings---totally discard the multi-way nature of data leading to a degradation of performance. 
	The first issue  one has to face when dealing with tensors is that the rank is not uniquely defined; many definitions are possibles, among which the canonical rank, Tucker rank and Tensor-Train are the most popular (\textit{cf} Section \ref{tensor algebra} for more details).
	Once  a suitable notion of rank has been established, a first approach to solve tensor completion consists in imposing a low-rank structure to the tensor to be recovered,  by leveraging, e.g. the Canonical Polyadic (CP)~\cite{Tom} or Tucker~\cite{Lat2} decomposition; the main drawback of these methods is that an estimate of the rank is required, which is not always available. A second approach relies on the assumption that matrix nuclear norm is a convex surrogate of rank, and tries to redefine an appropriate tensor nuclear norm \cite{Liu}, \cite{Gan}, \cite{Phien}, which generally leads to much reliable results.
	A somewhat hybrid path consists in imposing norms constraints on factorized matrices instead than on tensor unfolding, in order to reduce computational burden \cite{Ying} \cite{Liu2}. For a more comprehensive account on the topic of tensor factorization we refer the reader to \cite{Song}. 
	
	The idea of exploring implicit regularization in deep learning with the view of tensor factorization has been suggested and partially studied in the recent work of~\cite{Razin}, which only dealt with CP factorization. 
	In the present paper, we go further by addressing other two tensor decompositions, namely Tucker and Tensor-Train, which have proved themselves to be more expressive and manageable than CP.
	We define deep Tucker and Tensor-Train \textit{unconstrained} factorization and introduce the notion of tensor effective rank to characterize implicit regularization in these models.
	Our main contributions are:
	\begin{enumerate}[(i)]
	\itemsep=0.1cm
		\item We study tensor factorization in the overparameterized regime. More specifically, instead of constraining the rank of the tensor decomposition and reducing the number of parameters, we propose Tucker and Tensor-Train unconstrained factorization that have deep structure and representation flexibility.
		
		\item We generalize the notion of effective rank, introduced for matrices as a continuous surrogate of the matrix rank, to tensors and use it to characterize the implicit regularization in deep tensor factorization. To our knowledge, this is the first proposition on how to extend effective rank to the context of Tucker and Tensor-Train decompositions.
		
		\item We show experimentally that gradient descent on Tucker and Tensor-Train unconstrained factorization tends to produce solutions with low tensor rank. The rank depends on the type of factorization. As far as we are aware, this is the first time such behaviour has been analysed for Tucker and Tensor-Train tensors. 
		
		\item Our empirical study provides insights into the behaviour of tensor singular values in deep tensor factorization, and supports the observation that the implicit regularization has to be addressed from a dynamical perspective.

	\end{enumerate}

	\section{TENSOR ALGEBRA}\label{tensor algebra}
	
	In this section we introduce tensors and the related multi-linear algebra. For a more comprehensive account on this subject we refer the reader to \cite{Kolda}.
	We will denote vector by boldface lowercase letters, e.g. $\bm{v}\in\mathbb{R}^{I_1}$ whereas matrices and tensors will be denoted respectively by boldface uppercase  and calligraphic letters, e.g. $\bm{M}\in\mathbb{R}^{I_1\times I_2}$ and $\mathcal{T}\in\mathbb{R}^{I_1\times I_2\times \cdots\times I_N}$. 
	The $(i_1, i_2, \ldots, i_N)$-th entry of $\mathcal{T}\in\mathbb{R}^{I_1\times I_2\times \cdots\times I_N}$ will be denoted by $\mathcal{T}_{i_1, i_2, \ldots, i_N}$ where $i_n=1, 2, \ldots, I_n$, for all $n=[1,\ldots, N]$. Given two tensors $\mathcal{T}, \mathcal{S}\in \mathbb{R}^{I_1\times I_2\times \cdots\times I_N}$ their scalar product writes
	\begin{equation*}
		\langle\mathcal{T}, \mathcal{S} \rangle = \sum_{i_1}\sum_{i_2}\cdots \sum_{i_N}\mathcal{S}_{i_1, i_2\cdots i_N} \mathcal{T}_{i_1, i_2\cdots i_N},
	\end{equation*}
	and the Frobenius norm of $\mathcal{T}$ is defined as $\|\mathcal{T}\|_F=\sqrt{\langle\mathcal{T}, \mathcal{T}\rangle}$.
	Given a $3$-rd order tensor $\mathcal{T}\in\mathbb{R}^{I_1 \times I_2\times I_3}$, its slices are the matrices obtained by fixing one index: as an example, $\bm{T}[i_2]\in\mathbb{R}^{I_1\times I_3}$ is the is the slice obtained by fixing $I_2=i_2$.
	
	Given $N$ vectors $\bm{v}_1\in\mathbb{R}^{I_1}, \bm{v}_2\in\mathbb{R}^{I_2}, \ldots, \bm{v}_N\in\mathbb{R}^{I_N} $, their outer product is the tensor whose $(i_1, \ldots, i_N)$-th entry writes $\left(\bm{v}_1\circ \bm{v}_2\circ\cdots\circ \bm{v}_N\right)_{i_1, i_2\ldots, i_N}=(\bm{v}_1)_{i_1}(\bm{v}_2)_{i_2}\ldots (\bm{v}_N)_{i_N}$ for all $i_n\in[1,\ldots, I_n]$, $n\in[1, \ldots, N]$. An $N$-th order tensor $\mathcal{T}$ is called a rank-$1$ tensor if it can be written as the outer product of $N$ vectors, i.e. $\mathcal{T}=\bm{v}_1\circ \bm{v}_2\circ\cdots\circ \bm{v}_N$. This lead us to the first definition of tensor rank.
	\begin{dfn}
		The canonical rank of an arbitrary $N$-th order tensor $\mathcal{T}$ is the minimal number of rank-$1$ tensors that yield $\mathcal{T}$ in a linear combination.
	\end{dfn}
	The canonical rank is the most straightforward extension of the matrix rank to the tensor realm; however its computation is an NP-hard problem \cite{Has}, thus alternative and more maneuverable definitions of the tensor rank have been established. To define them let us define the (Tucker) $n$-unfolding of a tensor as follows:
	Given an $N$-th order tensor $\mathcal{T}\in\mathbb{R}^{I_1\times I_2\times \cdots\times I_N}$, the matrix  $\bm{T}_{(n)}\in\mathbb{R}^{I_n\times(I_{n+1}I_{n+2}\ldots I_NI_1I_2\ldots I_{n-1})}$ is called the (Tucker) unfolding of $\mathcal{T}$ along the $n$-th mode. This bring us to define the $n$-rank and the Tucker rank.
	\begin{dfn}
		The tensor $n$-rank
		of $\mathcal{T}$ is defined as the rank of $\bm{T}_{(n)}$. The Tucker rank, also called multi-linear rank, is defined as the $N$-tuple whose $i$-th entry is the $i$-th rank of $\mathcal{T}$.
	\end{dfn}
	Different $n$-ranks of a tensor are not necessarily the same and generally speaking they do not coincide with the canonical rank, though it can be found that any $n$-rank is bounded by the canonical rank. Tucker rank is intimately related to the higher-order singular value decomposition, on which our analysis relies. The first step to take is to define a proper way to multiply a tensor and a matrix.
	
	\begin{dfn}
		The $n$-mode product of a tensor $\mathcal{T}\in\mathbb{R}^{I_1\times I_2\times\cdots\times I_N}$ with a matrix $\bm{U}\in\mathbb{R}^{J\times I_n}$ is a tensor of size $(I_1\times I_2\times\cdots I_{n-1}\times J\times I_{n+1}\cdots\times I_N)$ denoted by $\mathcal{T}\times_{n}\bm{U}$
		and its entries are defined as
		\begin{equation*}
			\left(\mathcal{T}\times_{n}\bm{U}\right)_{i_1\cdots i_{n-1} j 
				i_{n+1}\cdots i_N}= \sum_{i_n=1}^{I_n}\mathcal{T}_{i_1\cdots i_{n-1} i_ni_{n+1}\cdots i_N}U_{j i_n}.
		\end{equation*}
	\end{dfn}
	The $n$-mode product satisfies to the following property: given $\mathcal{T}\in\mathbb{R}^{I_1\times I_2\times\cdots\times I_N}$ and two  matrices $\bm{F}\in\mathbb{R}^{J\times I_n}$, $\bm{G}\in\mathbb{R}^{K\times J}$, one has
	\begin{equation*}
		\left(\mathcal{T}\times_n\bm{F}\right)\times_n\bm{G} = \mathcal{T}\times_n\left(\bm{G}\cdot\bm{F}\right),
	\end{equation*}
	which generalizes to more than two matrices. We are in shape to write a Singular Value Decomposition~(SVD) applying to tensors, named Higher-Order Singular Value Decomposition (HOSVD).
	\begin{thm}[HOSVD, \cite{Lat}]
		Any tensor $\mathcal{T}\in\mathbb{R}^{I_1\times I_2\times\cdots\times I_N}$ can be written as the product
		\begin{equation*}
			\mathcal{T} = \mathcal{C}\times_1\bm{U}^{(1)}\times_2\bm{U}^{(2)}\cdots\times_N\bm{U}^{(N)},
		\end{equation*}
		where
		\begin{enumerate}
			\item $\bm{U}^{(n)}$ is a unitary $(I_n\times I_n)$-matrix, usually denoted as factor matrix;
			\item $\mathcal{C}$ is a $(I_1\times I_2\times\cdots I_N)$-tensor, usually referred to as core tensor, of which the sub-tensors $\mathcal{C}_{i_n=\alpha}$ obtained by fixing the $n$-th index to $\alpha$ are such that
			\begin{enumerate}[(i)]
				\item they are mutually orthogonal: given two sub-tensors $\mathcal{C}_{i_n=\alpha}$ and $\mathcal{C}_{i_n=\beta}$, they are orthogonal for all possible values of $n, \alpha, \beta$ subject to $\alpha\neq\beta$:
				\begin{equation*}
					\langle\mathcal{C}_{i_n=\alpha}, \mathcal{C}_{i_n=\beta} \rangle = 0,
				\end{equation*}
				\item they are ordered:
				\begin{equation*}
					\|\mathcal{C}_{i_n=1}\|_F\geq \|\mathcal{C}_{i_n=2}\|_F\cdots\geq \|\mathcal{C}_{i_n=I_n}\|_F\geq 0.
				\end{equation*} 
			\end{enumerate}	
		\end{enumerate}
	\end{thm}
	The analogy with matrix SVD is straightforward: the Frobenius norm of subtensors 
	$\|\mathcal{C}_{i_n=i}\|_F:=\sigma^n_i$ are the $n$-mode singular values and the column vectors of the matrices $\bm{U}^{(n)}$ are the $n$-mode singular vectors. As in the matrix case, where the number of non-zero singular values controls the rank, in the higher-order setup we have that, if $R_n$ is equal to the highest index for which $\|\mathcal{C}_{i_n=R_n}\|>0$ then one has that the $n$-rank of $\mathcal{T}$ is equal to $R_n$. Tucker decomposition \cite{Tuc} with rank $(R_1, R_2, \ldots, R_N)$ can be obtained from HOSVD by considering only the first $R_n$ column vectors of the matrix $\bm{U}^{(n)}$ for each $n\in[1,\ldots, N]$.
	
	\smallskip
	Tensor-Train (TT) decomposition \cite{Ose} writes $\mathcal{T}$ as a product of third order tensors \break $\mathcal{T}_k\in\mathbb{R}^{R_{k-1}\times I_k\times R_k}$ called cores. The $TT$-rank of the decomposition is given by the tuple $(R_0, R_1, \ldots, R_N)$ \break where $R_0=R_N=1$. By denoting $\bm{T}_k[i_k]$ the slice obtained from $\mathcal{T}_k$ by taking $I_k=i_k$, TT decomposition writes
	\begin{equation}
		\mathcal{T}_{i_1,i_2, \ldots i_N} = \bm{T}_1[i_1]\bm{T}_2[i_2]\ldots\bm{T}_N[i_N].
	\end{equation}
	TT algorithm is naturally associated with the following unfolding:
	given an $N$-th order tensor $\mathcal{T}\in\mathbb{R}^{I_1\times I_2\times \cdots\times I_N}$, the matrix  $\bm{T}_{[n]}\in\mathbb{R}^{(I_1 I_2\cdots I_n)\times(I_{n+1}I_{n+2}\ldots I_N)}$ is called the TT unfolding of $\mathcal{T}$ along the $n$-th mode.
	In order to define TT generalized singular values, we recall  TT-SVD algorithm~(\cite{Ose}, Algorithm 1).
	Given $\mathcal{T}\in\mathbb{R}^{I_1\times I_2\times\cdots\times I_N}$ and a TT-rank $(R_0, R_1,\ldots, R_N)$,
	it starts by considering the unfolding $\bm{T}_{[1]}$ and   its partial SVD denoted by $U_nS_nV_n^{T}$ truncated to the first $R_1$ singular values. Then, the first core $\mathcal{T}_{1}$ will be given by reshaping $U_1$ as a $3$-rd order tensor with size $(R_{0}, I_1, R_1)$ and, at the following iteration an SVD truncated to the first $R_!$ singular values is applied to $S_1V_1^{T}$. The algorithm stops after $N-1$ iterations. For any $n\in [1, \ldots, N-1]$, we say that the diagonal entries of $S_n$ are the $n$-mode singular values associated with TT decomposition.

	\section{DEEP~TENSOR~FACTORIZATION}
	\label{ufsection}
	In this section, we introduce Tucker and TT unconstrained factorization for deep tensor completion.
	Tensor completion by mean of tensor factorization is the generalization of Eq. \eqref{matrix_completion_loss}. Let $\mathcal{M}\in\mathbb{R}^{I_1\times I_2\times\cdots\times I_N}$ the ground truth tensor we want to recover and let us denote by $\Omega\subset\{1, 2, \ldots I_1\}\times\{1, 2,\ldots, I_2\}\times \cdots \times\{1, 2, \ldots, I_N\}$ the set which indexes the observed entries. We want to minimize the loss $\ell:\mathbb{R}^{I_1\times I_2\times\cdots\times I_N}\to\mathbb{R}_{\geq 0}$ defined as
	\begin{equation}\label{tensor_completion_loss}
		\ell(\mathcal{W}) = \frac{1}{2}\sum_{(i_1, \ldots, i_N)\in\Omega}\left(\mathcal{W}_{i_1, \ldots, i_N} - \mathcal{M}_{i_1, \ldots, i_N}\right)^2.
	\end{equation}
	To do so, we adopt a deep learning perspective and propose Tucker and TT \enquote{overparameterized} factorization where  no assumptions are made with respect to the rank. This is crucial in order to analyze the effect of the implicit regularization on deep tensor factorization.
	
	\paragraph{Deep Tucker Unconstrained Factorization}
	We write $\mathcal{W}$ as the product of a core tensor times a set of factor matrices and inspired by deep matrix factorization \cite{Arora}, we include depth by writing
	\begin{equation}
		\begin{split}\label{utf}
			\mathcal{W} =  \mathcal{D} & \times_1\left(\bm{V}^{(1)}_1\cdot \bm{V}^{(1)}_2 \cdots \bm{V}^{(1)}_{k_1}\right) \\
			& \times_2\left(\bm{V}^{(2)}_1\cdot \bm{V}^{(2)}_2 \cdots \bm{V}^{(2)}_{k_2}\right)\\
			& \times_3 \cdots \times_N \left(\bm{V}^{(N)}_1\cdot \bm{V}^{(N)}_2 \cdots \bm{V}^{(N)}_{k_N}\right),
		\end{split}
	\end{equation}
	where $\mathcal{D}$ is a $(I_1\times I_2\times\cdots\times I_N)$-th tensor and the matrices $\bm{V}^{(n)}_{i}$ are in $I_n\times I_n$ for all $n\in\{1,\ldots, N\}$ and $i\in\{1,\ldots,  k_n\}$
	and  we refer to Eq. \eqref{utf} as Tucker Unconstrained Factorization~(UF). A Tucker UF with $k_n$ factor matrices along mode $n$, $n\in[1,\ldots, N]$ will be said to have depth $(k_1, k_2, \ldots, k_N)$ and we adopt the convention that if $k_n=0$ for some $n$, then the $n$-mode multiplication writes $\mathcal{D}\times_n\mathds{1}_{I_n}$, where $\mathds{1}_{I_n}$ is the $I_n\times I_n$ identity matrix. Thus the depth-$(0, 0, \ldots, 0)$ Tucker UF writes
	\begin{equation*}
		\mathcal{W} = \mathcal{D}\times_1 \mathds{1}_{I_1}\times_2\mathds{1}_{I_2}\cdots, \times_N\mathds{1}_{I_N} = \mathcal{D}.
	\end{equation*}
	Notice that Eq. \eqref{utf} has no explicit constraints on any of the $n$-rank and that it 
	is the most natural way to carry over the deep matrix factorization setup addressed in \cite{Arora} to Tucker low-ranked tensors.
	
	\paragraph{Deep TT Unconstrained Factorization}
	As far as Tensor-Train decomposition is concerned, in order not to introduce any explicit constraint on the TT-rank (recall the TT-SVD algorithm in the previous section) we just need to keep the entire SVD of each $\bm{T}_{[n]}$; thus the unconstrained Tensor Train factorization of $\mathcal{W}$ writes:
	\begin{equation}\label{uttf}
		\mathcal{W}_{i_1, i_2,\ldots, i_n} = \bm{W}_1[i_1]\bm{W}_2[i_2]\ldots\bm{W}_N[i_N],
	\end{equation}
	such that, for each $n\in[1, \ldots, N-1]$, $\mathcal{W}_n\in\mathbb{R}^{R_{n-1}\times I_n\times  R_n}, \quad R_n = \min\left\{I_1I_2 \ldots I_n,  I_{n+1}\ldots I_N\right\}$ and with $R_0 = R_N = 1$. We denote Eq. \eqref{uttf} as Tensor-Train UF.
	The notion of depth here is intrinsic to the TT decomposition, since the number of core tensors is fixed according to the order of the original tensor.
	It is, however, possible to include other notions of depth in the parameterization given by Eq.~\eqref{uttf}. At least, two options can be explored: the first one is straightforward from Tucker UF and  consists in factorizing each TT core by means of Eq.~\eqref{utf}. The second strategy consists in writing each core $\mathcal{W}_n$, $ n\in[1, \ldots, N]$ by mean of a TT decomposition. 
	We defer this to future work.

	\section{TENSOR EFFECTIVE RANKS}\label{nuc_eff}
	
	When performing tensor completion, brute performance of the completion task is grasped by the reconstruction error, defined as the normalized Frobenius distance
	$\|\mathcal{W}^* -\mathcal{M}\|_F/\|\mathcal{M}\|_F$
	between the solution $\mathcal{W}^*$  of Eq. \eqref{tensor_completion_loss} and the ground-truth tensor $\mathcal{M}$.\par As far as Tucker factorization is concerned, a suitable extension of the nuclear norm to the tensor realm has been proposed by \cite{Liu}, and writes
	\begin{equation*}
		\|\mathcal{W}\|_*^{tuc} = \sum_{n=1}^N \alpha_n\|\bm{W}_{(n)}\|_* \quad\text{s.t.}\quad \sum_{n=1}^{N}\alpha_n = 1.
	\end{equation*}
	In the aforementioned paper, authors also established several algorithms based on the minimization of $\|\cdot\|_*^{tuc}$, the most effective of which is named High Accuracy Low Rank Tensor Completion (HALRTC) 
	and takes advantage of Alternating Direction Method of Multipliers (ADMM) to perform the task. 
	On the other hand, when dealing with TT factorization, \cite{Phien} introduced a convenient notion of nuclear norm, which reads
	\begin{equation*}
		\|\mathcal{W}\|_*^{TT} = \sum_{n=1}^{N-1} \alpha_n\|\bm{W}_{[n]}\|_* \quad\text{s.t.}\quad \sum_{n=1}^{N-1}\alpha_n = 1,
	\end{equation*}
	as well as an algorithm, named TT-SILRTC, to deal with the minimization of $\|\cdot\|^{TT}_*$ by means of Block Coordinate Descent method.
	These norms arise as convex approximations of the rank functions which are much easier to handle analytically and numerically.

	In the matrix case, the notion of effective rank has also been proposed in \cite{Roy} as a continuous surrogate of the matrix rank and it proved itself, as remarked in~\cite{Arora}, to provide useful insights for the deep matrix factorization setup.
	The effective rank of a matrix is computed from the entropy of its singular value distribution.
	In this respect we propose a new definition of effective rank which allow us to cover Tucker and TT ranks. We start by recalling the definition of Shannon entropy: given  a discrete probability distribution $\{p_i\}_{i=1}^n$, the Shannon entropy $H\left(\{p_i\}_{i=1}^n\right)$ is defined as $H\left(\{p_i\}_{i=1}^n\right)= - \sum_i p_i\cdot \ln p_i$, with the convention that $0\cdot\ln 0 =0$.
	
	\smallskip
	\begin{dfn}[Tucker effective rank]
		Let $\mathcal{T}\in\mathbb{R}^{I_1\times I_2\times\cdots\times I_N}$ be a $N$-th order tensor and let $\mathcal{C}\times_1\bm{U}^{(1)}\times_2\bm{U}^{(2)}\cdots\times_N\bm{U}^{(N)}$ be the associated HOSVD. For $n\in[1, \ldots N]$, consider the $n$-mode singular values $\{\sigma^n_i\}_{i=1}^{I_n}$ and the associate probability distribution $\left\{\rho^n_i = \sigma^n_i(\mathcal{T}) / \sum_{j=1}^{I_n}\sigma^n_j\right\}_{i=1}^{I_n}$.
		The Tucker effective rank along the $n$-th mode of $\mathcal{T}$ is defined as $erank^{tuc}_n(\mathcal{T}) = \exp \left\{H(\rho^n_1(\mathcal{T}), \rho^n_2(\mathcal{T}), \ldots, \rho^n_{I_n}(\mathcal{T}))\right\}$.
	\end{dfn} 
	
	\smallskip
	\begin{dfn}[TT effective rank]
		Let $\mathcal{T}\in\mathbb{R}^{I_1\times I_2\times\cdots\times I_N}$ and let $\bm{U}_n\bm{S}_n\bm{V}_n^T$ be the SVD associated with the $n$-th unfolding $\bm{T}_{[n]}$ given by the TT-SVD algorithm, denote by $\{\sigma^n_i\}_{i=1}^{I_n}$ the singular values and consider the associate probability distribution $\left\{\rho^n_i = \sigma^n_i(\mathcal{T}) / \sum_{j=1}^{I_n}\sigma^n_j\right\}_{i=1}^{I_n}$.
		The TT effective rank along the $n$-th mode of $\mathcal{T}$ is defined as \break $erank^{TT}_n(\mathcal{T}) = \exp\left\{H(\rho^n_1(\mathcal{T}), \rho^n_2(\mathcal{T}), \ldots, \rho^n_{I_n}(\mathcal{T}))\right\}$.
	\end{dfn}
	It is easy to see that Proposition $1$ in \cite{Roy}, which gives lower and upper bounds of matrix effective rank, can be straightforwardly extended to cover Tucker and TT effective ranks.

	\begin{property}
		Let $R_n$ be the Tucker $n$-rank (respectively TT $n$-rank), we have that the Tucker effective rank (resp. TT effective rank) is bounded by
		$1\leq erank_n^{tuc}(\mathcal{T})\leq R_n$ (resp. $1\leq erank_n^{TT}(\mathcal{T})\leq R_n$).
	\end{property}
	
	As for the case of matrix completion, a crucial question, which may have important implications in understanding the implicit regularization in deep learning, is whether gradient descent in deep tensor factorization converges to solution with minimal nuclear norm, or whether there are other measures that are more relevant to characterize this phenomenon.
	The experiments described in the next section were designed to provide some clues to this question.

\begin{figure*}
\centering
\begin{subfigure}{.5\textwidth}
  \centering
  \includegraphics[scale=0.5]{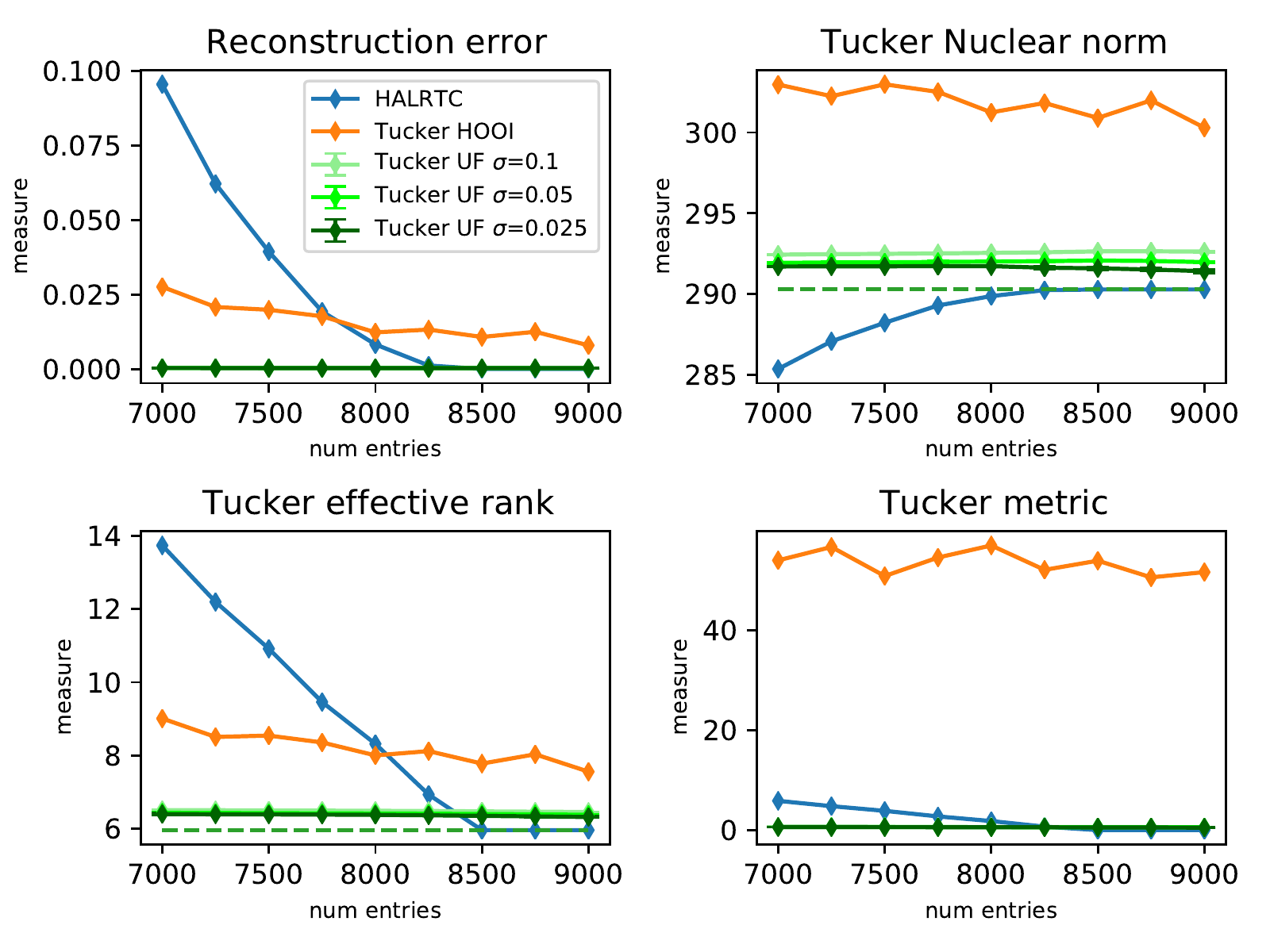}
  \caption{}
  \label{fig:sub1}
\end{subfigure}%
\begin{subfigure}{.5\textwidth}
  \centering
  \includegraphics[scale=0.5]{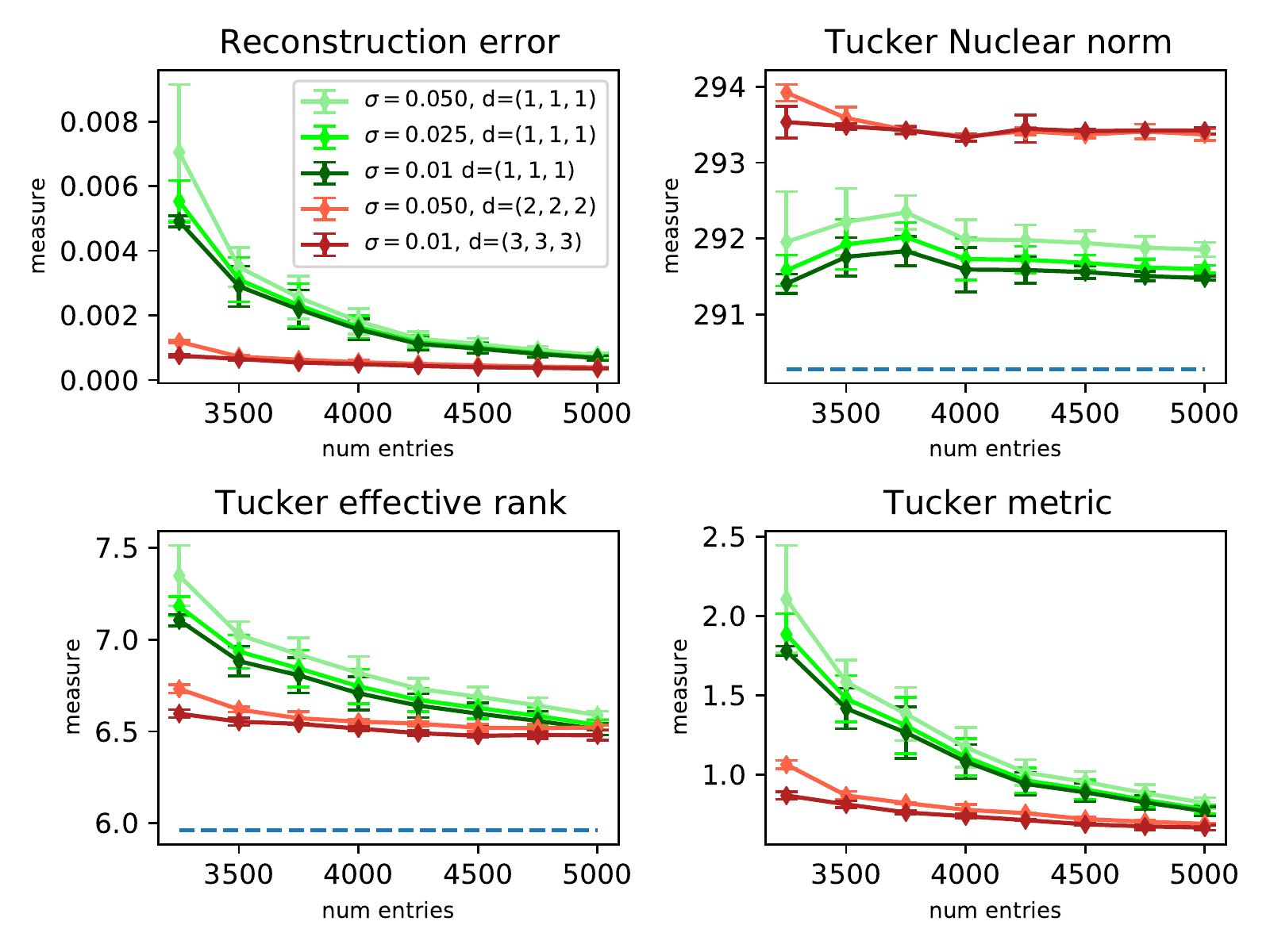}
  \caption{}
  \label{fig:sub2}
\end{subfigure}
\caption{Tensor completion experiments of a $30\times 30 \times 30$ tensor with Tucker rank equal to $(6, 6, 6)$ by using Tucker UFs. Dashed lines represent ground truth values of the observable considered. Figure $(a)$ compares depth ~$(1, 1, 1)$ Tucker UF with HALRTC and Tucker HOOI (where Tucker rank has been set to $(8, 8, 8)$) methods. In~$(b)$ we compare Tucker UFs with different depth from $(1,1,1)$ to $(3,3,3)$.}\label{synthetic_tucker}
\end{figure*}
\begin{figure*}
\begin{center}
   \includegraphics[scale=0.55]{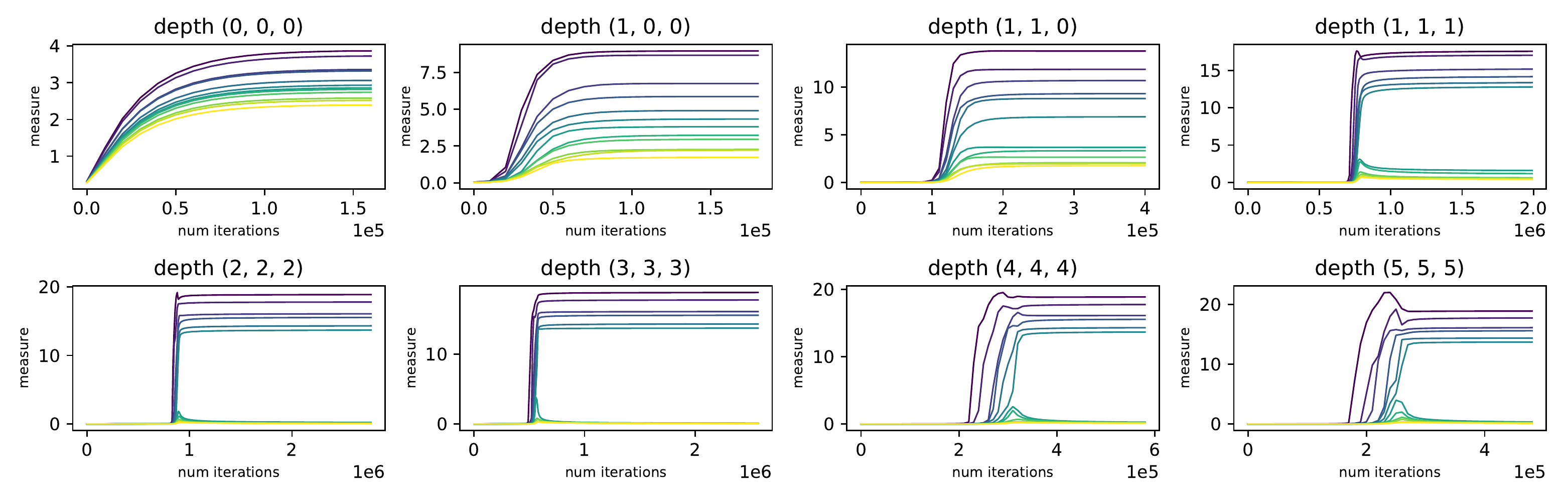}
\end{center}
\caption{Time evolution of singular values  over iterations of SGD training process to perform tensor completion of a $30\times 30 \times 30$ tensor with Tucker rank equal to $(6, 6, 6)$ by using Tucker UFs with different depth
for Tucker factorization. In each experiment $2750$ entries are observed. Plots represents the $12$ largest $2$-mode singular values.}
\label{singular_values_tucker}
\end{figure*}



\begin{figure*}
\centering
\begin{subfigure}{.5\textwidth}
  \centering
  \includegraphics[scale=0.45]{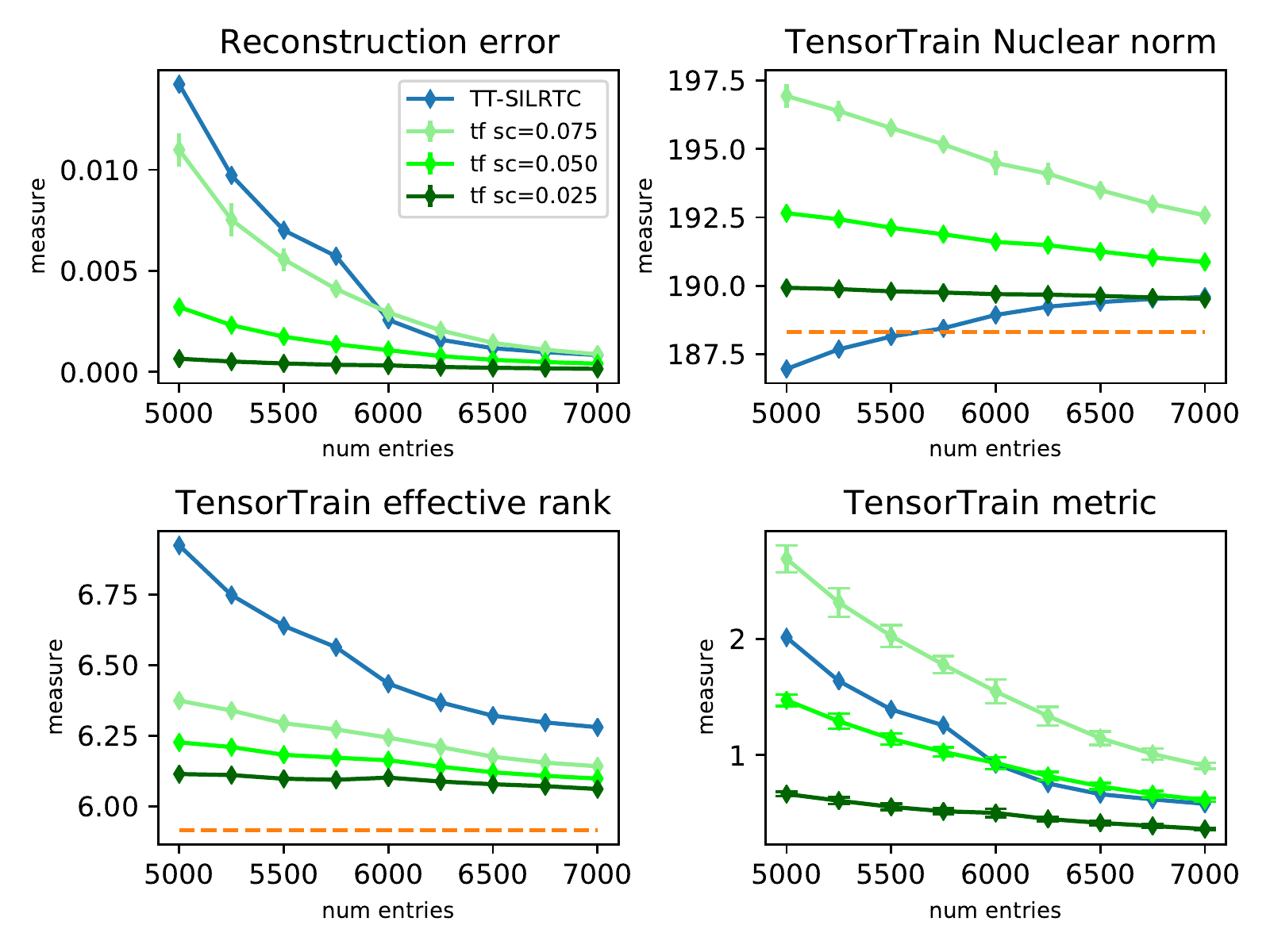}
  \caption{}
  \label{fig:sub1}
\end{subfigure}%
\begin{subfigure}{.6\textwidth}
  \centering
  \includegraphics[scale=0.6]{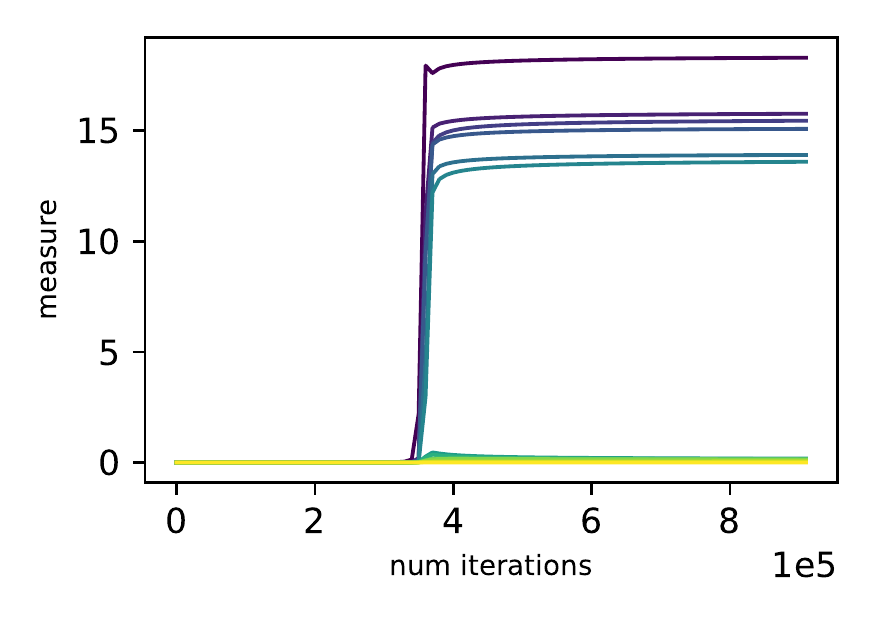}
  \caption{}
  \label{fig:sub2}
\end{subfigure}
\caption{Tensor completion experiments on a $12 \times 12 \times 12 \times 12$ tensor with Tensor-Train rank equal to $(1, 6, 6, 6, 1)$. Dashed lines represent ground truth values of the observable considered. Figure $(a)$ compares TT UFs with TT-SILRTC whereas in $(b)$ the $12$ largest $2$-mode singular values are depicted.}
\label{synthetic_TT}
\end{figure*}



\begin{figure*}
\centering
\begin{subfigure}{.5\textwidth}
  \centering
  \includegraphics[scale=0.55]{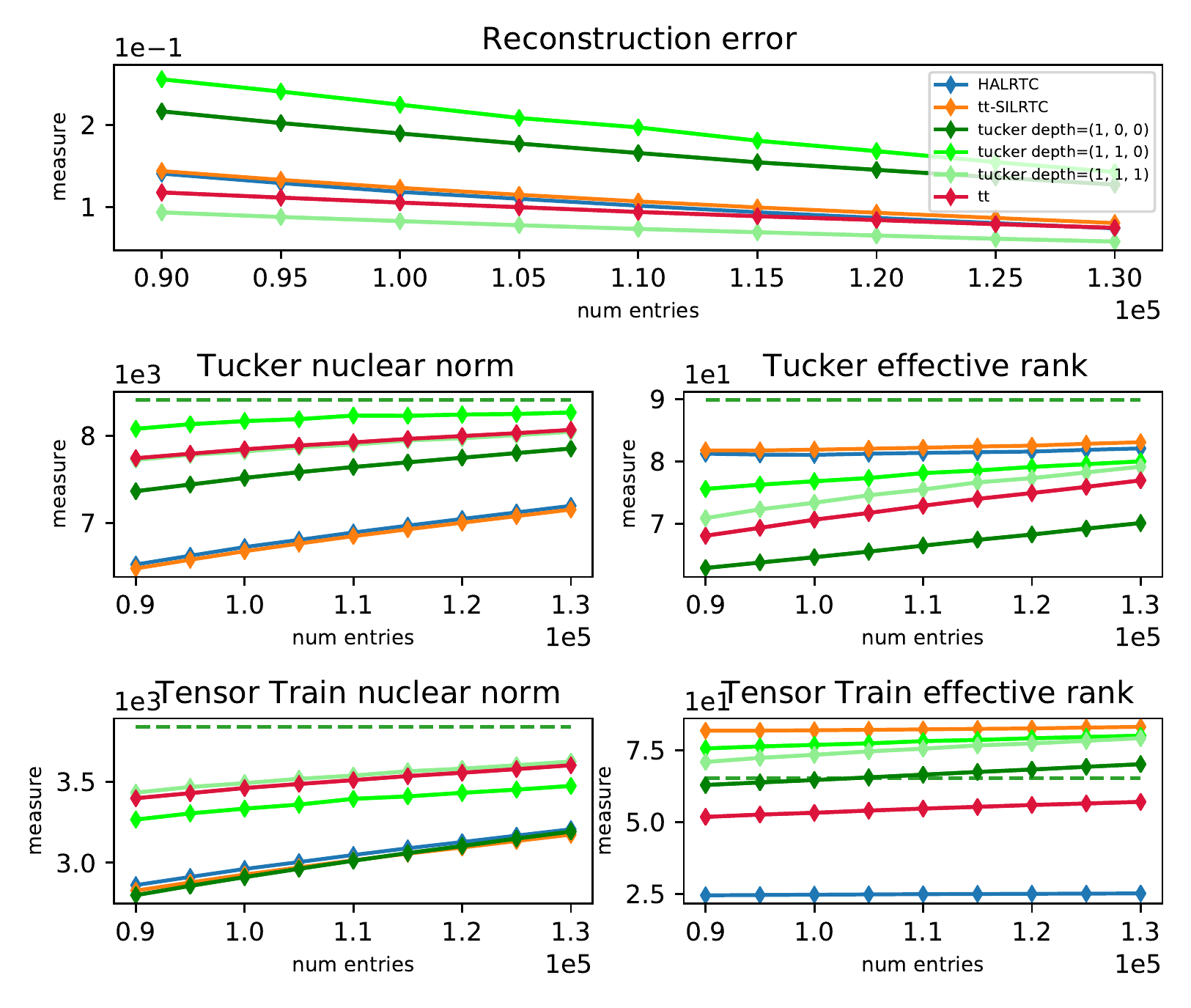}
  \caption{}
  \label{fig:sub1}
\end{subfigure}%
\begin{subfigure}{.5\textwidth}
  \centering
  \includegraphics[scale=0.55]{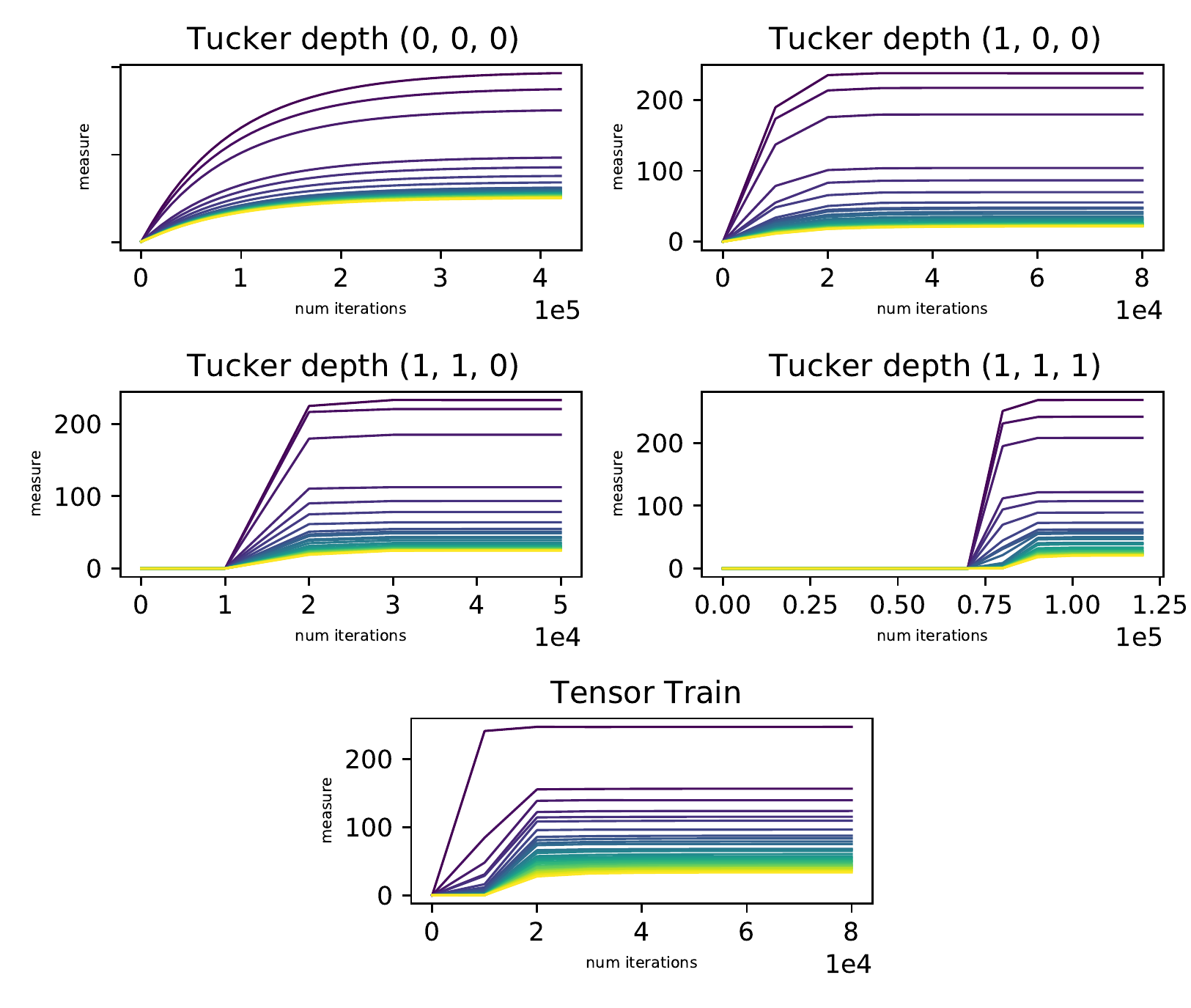}
  \caption{}
  \label{fig:sub2}
\end{subfigure}
\caption{Tensor completion experiments on the CCDS data set. $(a)$ shows a comparison between Tucker and TT UFs together with HALRTC and TT-SILRTC. 
$(b)$ shows time evolution of $2$-mode singular values where $90000$ entries were observed.}
\label{all_CCDS}
\end{figure*}


	\section{EXPERIMENTAL ANALYSIS}
	
	Our tensor completion experiments focus on the most interesting regime where few entries are observed. They are designed to better understand the implicit regularization in deep tensor factorization. We first evaluate the reconstruction performance of Tucker and TT (unconstrained) factorization. We then explore whether the implicit regularization induced by gradient-based optimization can be characterized by the minimization of tensor norms or of tensor ranks. Finally, we investigate the impact of the implicit regularization on the singular values dynamics of the learned tensor. 

	\subsection{Experimental setup}

	\textbf{Synthetic data.} We generated a $30\times 30\times 30$ tensor with Tucker-rank equal to $(6, 6, 6)$ and entries sampled from a centered and reduced normal distribution. Recalling Eq. \eqref{utf}, entries of $\mathcal{W}$ are sampled from a centered normal distribution with  variance proportional to $\sigma$,  
$\sigma\in\{0.1, 0.05, 0.025\}$.
We also generated a $4$-th order tensor of size $12\times 12\times 12\times 12$ with TT-rank equal to $(1, 6, 6, 6, 1)$ with entries sampled from a centered normal distribution with variance proportional to $\sigma$, 
$\sigma \in\{0.075, 0.05, 0.025\}$.\par

\textbf{Real~data.} The comprehensive climate data set~(CCDS) is a collection of climate records of North America \cite{Loz}. The data set contains monthly observations of $17$ variables such as carbon dioxide and temperature spanning from $1990$ to $2001$ across $125$ observation locations thus giving a $138\times 17\times 125$ tensor.\footnote{The aggregated and processed data set can be found at https://viterbi-web.usc.edu/~liu32/data.html.} The Meteo-UK data set is collected from the meteorological office in UK.\footnote{http://www.metoffice.gov.uk/public/weather/climate-historic} It contains monthly measurements of $5$ variables in $16$ stations across UK from $1960$ to $2000$ resulting in a $492\times 5\times 16$ tensor; measures were scaled to have zero mean and unitary variance.
\par

\textbf{Training.} 
Minimization of the loss given by Eq.~\eqref{tensor_completion_loss} is done by backpropagation using SGD on full batch; training is stopped when training loss is smaller than $10^{-6}$ or when $4\cdot 10^{6}$ iteration are elapsed. Learning rate is set to $0.1$. When measuring reconstruction error, nuclear norms and effective ranks, 
we took the average over three runs initialized with different seeds.\par

\textbf{Baselines.} All operations involving tensor decomposition are performed with TensorLy \cite{tensorly}. We compared Tucker UF to HALRTC algorithm~\cite{Liu} and to Tucker HOOI algorithm described in~\cite{Lat2}. 
Both these baseline are available in the PyTen Toolbox~\cite{Song}. As far as the TT decomposition is concerned, we compared TT UF with TT-SILRTC~\cite{Phien}.\par

\subsection{Reconstruction Error and Depth Effect}

\begin{figure*}
\centering
\begin{subfigure}{.5\textwidth}
  \centering
  \includegraphics[scale=0.5]{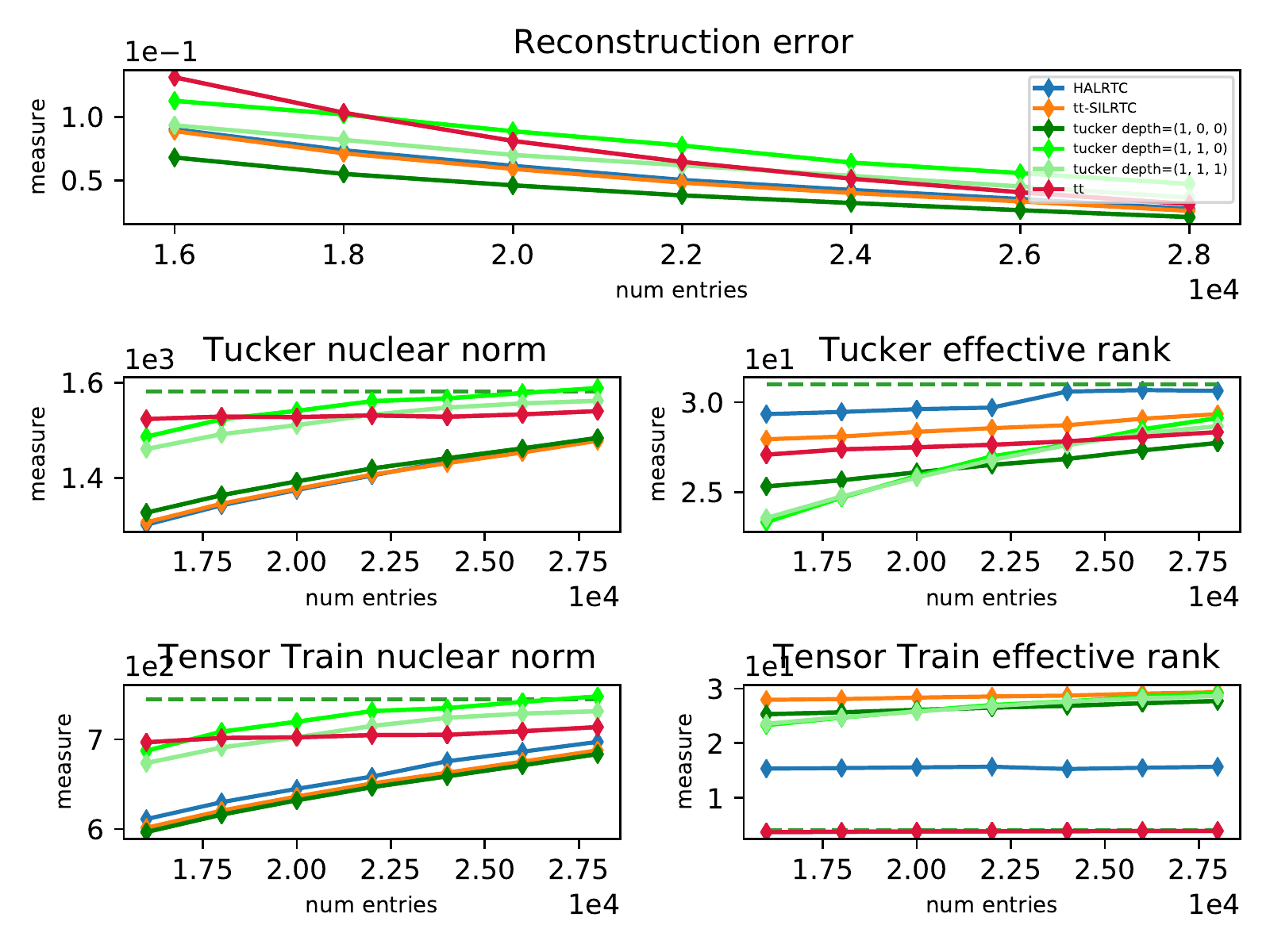}
  \caption{}
  \label{fig:sub1}
\end{subfigure}%
\begin{subfigure}{.5\textwidth}
  \centering
  \includegraphics[scale=0.5]{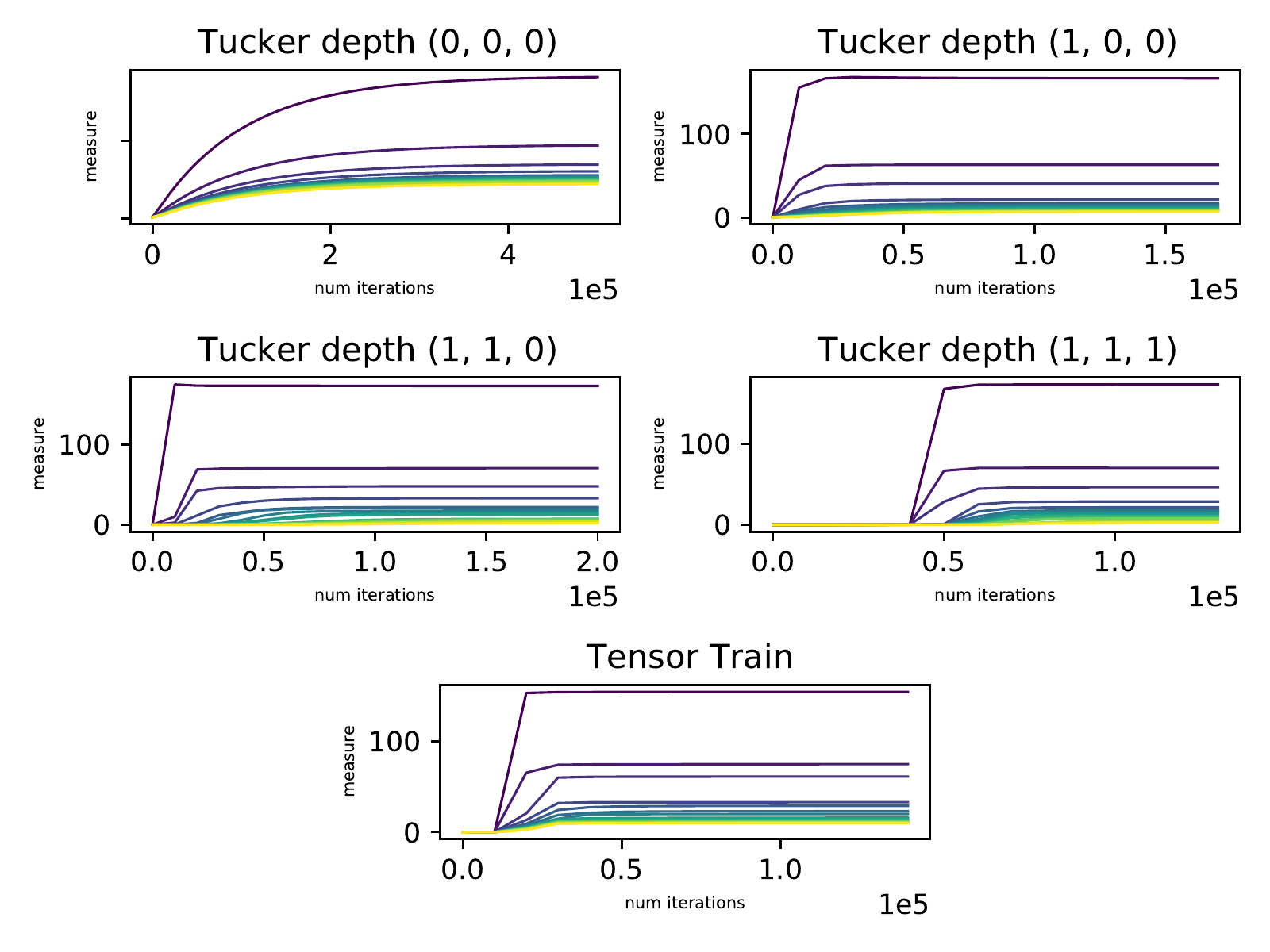}
  \caption{}
  \label{fig:sub2}
\end{subfigure}
\caption{Tensor completion experiments on the Meteo UK data set. $(a)$ shows a comparison between Tucker and TT UFs together with HALRTC and TT-SILRTC. 
$(b)$ shows time evolution of $2$-mode singular values singular values, where $20000$ entries were observed.}
\label{all_meteo_uk}
\end{figure*}

For the sake of readability, we point out that the label ``measure'' on the $y$-axis of all figures refers to the values taken by the observable specified in the plot's title. Let us start by looking at the ``reconstruction error'' plots in Figures
\ref{synthetic_tucker} and \ref{synthetic_TT} concerning experiments on synthetic tensors with low Tucker rank and TT rank respectively. We can see that both deep Tucker and TT UF outperforms nuclear norm based methods, in particular for the regime where fewer entries are observed (left side of the plots). We also observe that by reducing the variance of the normal distribution used at initialization recovery capacity increases as well. In the case of Tucker UF, we get by Figure \ref{synthetic_tucker} $(b)$ that depth plays a beneficial role on completion task. When considering the  reconstruction error on CCDS (Figure \ref{all_CCDS}(a)) and Meteo UK (Figure \ref{all_meteo_uk}(a)) we see that the tendency is much less noticeable with all method giving comparable results.


\subsection{Nuclear~Norm~vs.~Effective~Rank}

Let us now analyze ``Tucker nuclear norm'' and ``TT nuclear norm'' plots in Figures
\ref{synthetic_tucker} and \ref{synthetic_TT}(a). We notice that both Tucker and TT UF do not minimize nuclear norm as much as HALRTC and TT-SILRTC~(i.e. nuclear norm-based method) do:
HALRTC and TT-SILRTC curves lie indeed below Tucker and TT UF's. The situation is reversed as soon as we look at the ``Tucker effective rank'' and ``TT effective rank'' plots: they show that Tucker  and TT UFs  minimize effective rank much more than HALRTC and TT-SILRTC methods. This fact gives us strong insight about regularization associated with gradient descent, as it pushes us to questioning Conjecture \ref{con1}. Nuclear norms plots for real data in Figures \ref{all_CCDS}(a) and \ref{all_meteo_uk}(a) confirm the tendency we just described for synthetic data.
In the case of synthetic data where the rank is known, we also consider Tucker \enquote{metric} (resp. TT \enquote{metric}) which 
we define as the Frobenius norm of the difference between the ground truth tensor and a low-rank Tucker (resp. TT) decomposition of the tensor learned by solving the deep Tucker (resp. TT) UF.\footnote{Note that here we abusively use the term \enquote{metric} since all the properties of a distance may not be satisfied.} 
For these two metrics UFs outperform nuclear norms based methods (see Figures \ref{synthetic_tucker} and \ref{synthetic_TT}(a)).\par

\subsection{Singular Value Dynamics}
Let us now examine plots displaying time trajectories of generalized singular values (Figures \ref{singular_values_tucker}, \ref{synthetic_TT}(b), \ref{all_CCDS}(b) and \ref{all_meteo_uk}(b)). In the case of Tucker UF we clearly see that the deeper the model, the sparser the solution given by SGD; moreover, for deep factorization, the transition from non-sparse to sparse solutions happens more abruptly. 
This observation is consistent with previous results in the matrix case~\cite{Arora}.
Sparsity can also be observed for TT UF, though in this case depth comparison was not possible. For the matrix case, Theorem 3 in \cite{Arora} gives a theoretical proof of the behavior of these plots. Roughly stated, it says that 
\begin{equation}
\label{eq:SVbehaviourMatrix}
\dot{\sigma}(t)\propto(\sigma(t))^{2(1-1/L)}\cdot \gamma(t),
\end{equation}
where $\gamma(t)$ is the reconstruction error at iteration $t$, and  $\sigma$ and $L$ being a singular value and the depth of the matrix factorization and it can be interpreted by saying that the exponent controlling the sparsity of the solution is the depth $L$. For deep Tucker UF, Figure \ref{singular_values_tucker} shows that depth plays a key role in driving the learned tensor toward a sparser solution; though experiments did not allow to isolate the behaviour of a single mode as depth $k_n$ along this mode varies, it is reasonable to suppose that, analogously to the matrix setup, $k_n$ might be indeed the exponent which controls sparsity.

\begin{figure}[t]
    \centering
    \includegraphics[scale=0.13]{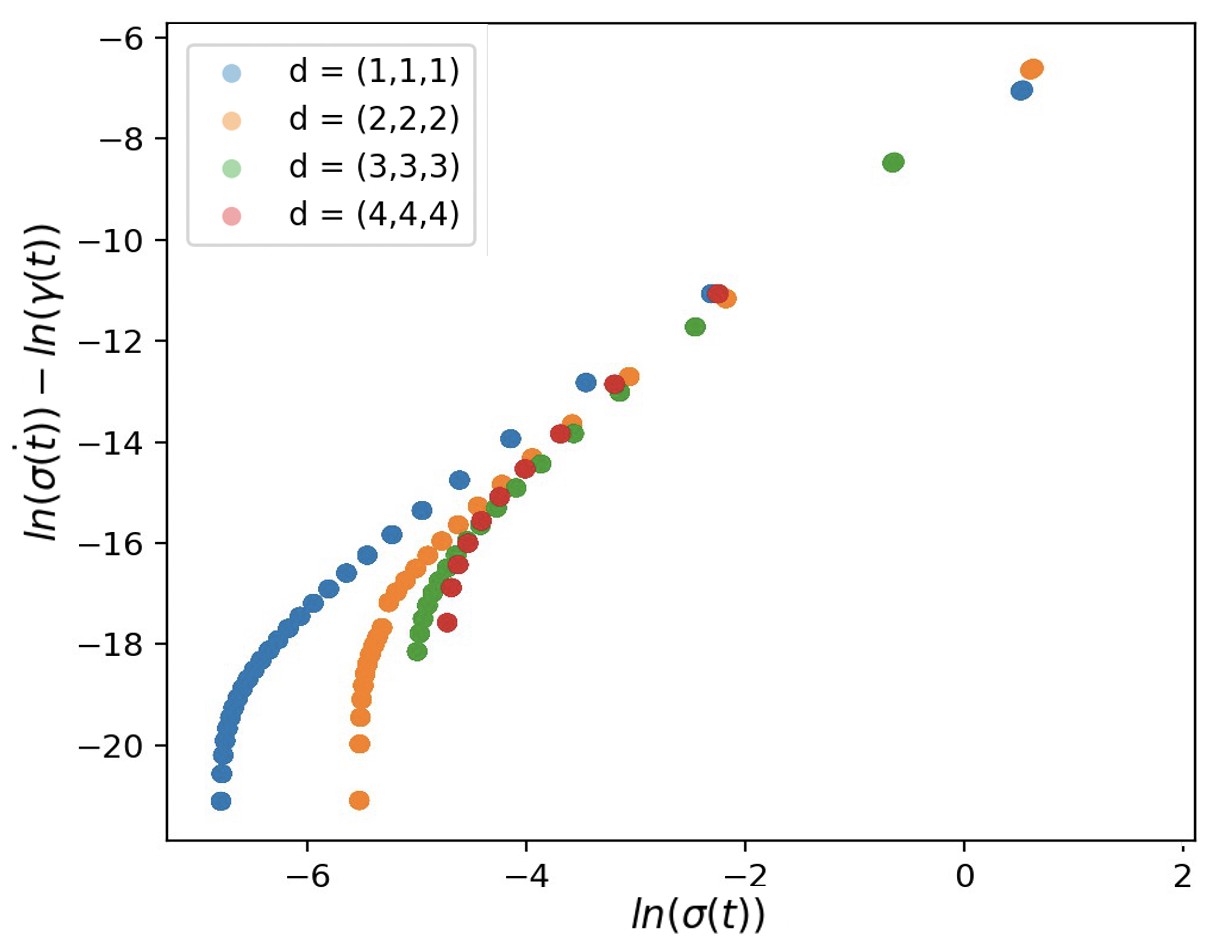}
    \caption{Behaviour of the singular value dynamics obtained from a $30\times 30 \times 30$ tensor with Tucker rank equal to $(6, 6, 6)$ and for depths ranging from $(1,1,1)$ to $(4,4,4)$. Lower left part stand for the beginning of the learning (high loss, small singular value with small derivatives). Upper right part corresponds to convergence (final singular values with  small derivatives, small loss). }
    \label{fig:SVbehaviour}
\end{figure}

We designed an experiment to investigate if a similar behaviour as the one for the matrix case could be observed in the tensor case. Figure~\ref{fig:SVbehaviour} shows the evolution of $\ln(\dot{\sigma}_i^n(t)) - \ln(\gamma(t))$ as a function of $\ln(\sigma_i^n(t))$ for different depth values. The linear behaviour of the curves gives an empirical argument in favour of the singular values dynamics provided in the case of matrix completion (see Eq.~\ref{eq:SVbehaviourMatrix}).
It is also interesting to note that, similarly to the matrix case, the slope of the curve rises with the depth of Tucker Unconstrained Factorization.

More formally, let us consider deep tensor factorization and a $(k_1, k_2, \ldots, k_N)$-Tucker UF. Analogously to the matrix case, one can  imagine that there exists a  function on the depths of all the tensor modes, $f$, so that the generalized singular values $\sigma_i^n$,  $\forall n\in[1, \ldots, N]$, $\forall i\in[1, \ldots, I_n]$, evolve according to \begin{equation}
\label{eq:SVbehaviour}
\dot{\sigma}_i^n(t)\propto(\sigma_i^n(t))^{f(k_1, \ldots, k_N)}\cdot \gamma(t).
\end{equation}

 \section{CONCLUSION}
 In this paper we explore the implicit regularization induced by gradient descent optimization in deep tensor factorization using the notion of effective rank. Our study provides empirical evidence that tensor nuclear norms is not minimized along the training process, which however tends to produce low tensor rank solutions. In the case of Tucker UF we were able to verify the beneficial effects of depth on the reconstruction error and to formulate a conjecture which gives account of the behavior of generalized singular values. Future research will focus on investigating deeper the dependency of the function $f$ that governs the dynamics of the singular values. 
	
	\bibliographystyle{IEEEtran}
	\bibliography{bibliography}

\end{document}